\documentclass[11pt,twoside]{article}

\usepackage{fullpage}

\usepackage{epsf}
\usepackage{fancyhdr}
\usepackage{graphics}
\usepackage{graphicx}
\usepackage{psfrag}
\usepackage{microtype}
\usepackage{subfigure}
\usepackage{algorithmic}

\usepackage[linesnumbered,ruled]{algorithm2e}
\DontPrintSemicolon
\usepackage{color}

\usepackage{amsthm}
\usepackage{amsfonts}
\usepackage{amsmath}
\usepackage{amssymb,bbm}
\usepackage{wrapfig}
\usepackage{tikz}
\usetikzlibrary{positioning}

\usepackage{tcolorbox}

\usepackage{natbib}

\usepackage{url}
\usepackage[colorlinks,linkcolor=magenta,citecolor=blue, pagebackref=true,backref=true]{hyperref}
\renewcommand*{\backrefalt}[4]{%
    \ifcase #1 \footnotesize{(Not cited.)}%
    \or        \footnotesize{(Cited on page~#2.)}%
    \else      \footnotesize{(Cited on pages~#2.)}%
    \fi}

\long\def\comment#1{}
\usepackage{nicefrac}
\usepackage[normalem]{ulem}
\usepackage{chngpage}

 \usepackage{tabularx}%

\usepackage{enumitem}
\usepackage{booktabs}
\usepackage{caption}

\usepackage{mathtools}

\usepackage{fullpage}

\newtheorem{theorem}{Theorem}[section]

\newtheorem{proposition}[theorem]{Proposition}

\newtheorem{remark}[theorem]{Remark}

\usepackage{bm}
\usepackage{makecell}
\usepackage{cancel}

\usepackage{bigints}

\newcommand{\policy}{\pi}
\newcommand{\policyy}{\pi^{'}}

\newcommand{\E}{\mathbb{E}}
\newcommand{\R}{\mathbb{R}}
\newcommand{\Pro}{\mathbb{P}}

\newcommand{\N}{\mathbb{N}}

\newcommand{\calA}{\mathcal{A}}

\newcommand{\calS}{\mathcal{S}}

\newcommand{\bI}{\mathbf{I}}

\newcommand{\bd}{\mathbf{d}}

\newcommand{\br}{\mathbf{r}}

\newcommand{\bv}{\mathbf{v}}

\newcommand{\bP}{\mathbf{P}}

\usepackage{tikz}
\usetikzlibrary{automata, positioning}
\usepackage{makecell}
\usepackage{graphpap,amscd,mathrsfs,graphicx,lscape,enumitem,dsfont,bm,url,subfigure}
\usepackage{epsfig,amstext,xspace}

\usepackage{thmtools,thm-restate}
\usetikzlibrary{calc}

\begin{document}

\begin{center}

{\bf{\LARGE{A General Perspective on Objectives of Reinforcement Learning}}}

\vspace*{.2in}
{\large{
\begin{tabular}{ccc}
Long Yang
 \end{tabular}
}}


\begin{tabular}{c}
School of Artificial Intelligence, Peking University, Beijing, China
\end{tabular}

\begin{tabular}{c}
\texttt{yanglong001@pku.edu.cn}
\end{tabular}

\vspace*{.2in}

\today

\vspace*{.2in}

\begin{abstract} 
In this lecture, we present a general perspective on reinforcement learning (RL) objectives, where we show three versions of objectives.
The first version is the standard definition of objective in RL literature. Then we extend the standard definition to the $\lambda$-return version, which unifies the standard definition of objective.
Finally, we propose a general objective that unifies the previous two versions.
The last version provides a high level to understand of RL's objective, where it shows a fundamental formulation that connects some widely used RL techniques (e.g., TD$(\lambda)$ and GAE), and this objective can be potentially applied to extensive RL algorithms.
\end{abstract}

\end{center}

\addtocounter{page}{0}
\thispagestyle{empty}

\setcounter{tocdepth}{2}
\tableofcontents

\addtocounter{page}{-1}
\thispagestyle{empty}
\newpage

\clearpage

\section{Introduction}

Although reinforcement learning (RL) is widely applied to extensive fields, there is stills lack a work that establishes the objective of starting from RL from the Markov decision process, which is very unfriendly to beginners.
To fill the gap in this view, in this lecture, we provide a self-contained, teachable technical introduction to the objectives of RL, where each section tackles a particular line of work from the transition probability matrix over the Markov decision process, reward, Bellman equation, discounted state distribution, and objectives.

Concretely, this lecture provides three equivalent versions of objectives.
The first version is presented in Theorem \ref{them:obj-01}, where it shows the objective as the expectation with respect to the random variable $(s,a,s^{'})$.
Theorem \ref{them:obj-01} illustrates all the random factors in the Markov decision process (MDP), and we refer to it as the \emph{standard objective} of MDP.
Furthermore, Theorem \ref{lem:lam-return-objective} extends and unifies the objective that appears in Theorem \ref{them:obj-01}.
Theorem \ref{lem:lam-return-objective} is traceable to TD$(\lambda)$ \citep{sutton1984temporal,sutton1988learning}, and we present it as the expectation with respect to the random variable the state $s$, where the state $s$ follows the $\lambda$-version of discounted state distribution.
Finally, we present a general objective that unifies the previous two versions (see Theorem \ref{objective-td-error-version}), which provides a high level to understand of RL's objective, where it shows a fundamental formulation that connects some widely used RL techniques (e.g., TD$(\lambda)$ and GAE), and this objective can be potentially applied to extensive RL algorithms.
For example, \cite{yang2022constrained} apply the main technique of Theorem \ref{objective-td-error-version} to obtain the surrogate function with respect to GAE \citep{schulman2016high}.
Although GAE has been widely used in RL, it lacks a theoretical analysis of the related algorithms.
Theorem \ref{objective-td-error-version} provides a possible way to establish GAE and empirical results by rigorous analysis.
To clarify this view, we present a surrogate function with respect to GAE, see Section \ref{application-gae}, where it provides a theoretical fundament for policy optimization with GAE.

\section{Markov Decision Process}

Reinforcement learning (RL) \citep{sutton2018reinforcement} is often formulated as 
a \emph{Markov decision process} (MDP) \citep{howard1960dynamic,puterman2014markov}. 
In this section, we review some necessary notation w.r.t. MDP.

An MDP is described as a tuple 
$
\mathcal{M}=(\mathcal{S},\mathcal{A},\mathbb{P},r,\rho_0,\gamma).
$
\begin{itemize}
\item $\mathcal{S}$ is the state space;
\item $\mathcal{A}$ is the action space;
\item $\Pro(\cdot|\cdot,\cdot):\calS\times\calA\times\calS\rightarrow[0,1]$, each $\mathbb{P}(s^{'}|s,a)$ denotes the probability  
of state transition from $s$ to $s^{'}$ underplaying the action $a$;
\item
$r(\cdot|\cdot,\cdot):\calS\times\calA\times\calS\rightarrow \R$; each $r(s^{'}|s,a)$ denotes the reward
of state transition from $s$ to $s^{'}$ underplaying the action $a$;
\item
$\rho_{0}(\cdot):\mathcal{S}\rightarrow[0,1]$ is the initial state distribution;
\item $\gamma\in(0,1)$ is the discount factor.
\end{itemize}
The probability and reward satisfy Markov property, i.e., $\mathbb{P}(s^{'}|s,a)$ and $r(s^{'}|s,a)$ only depend on the immediately preceding state $s$ and action $a$, not at all on earlier states and actions.

A stationary Markov policy $\pi$ is a probability distribution defined on $\mathcal{S}\times\mathcal{A}$, $\pi(a|s)$ denotes the probability of playing $a$ in state $s$.
We use $\Pi$ to denote the set that collects all the stationary Markov policies. 
Let 
\begin{flalign}
\tau=\{s_{t}, a_{t}, r_{t+1}\}_{t\ge0}\sim \pi
\end{flalign} 
be the trajectory generated by $\pi$, 
where 
\[s_{0}\sim\rho_{0}(\cdot),~a_{t}\sim\pi(\cdot|s_t),~s_{t+1}\sim \mathbb{P}(\cdot|s_{t},a_{t}),~\text{and}~r_{t+1}=r(s_{t+1}|s_t,a_t).\]

\subsection{Single-Step Transition Probability Matrix}
Let $\mathbf{P}_{\pi}\in\R^{|\calS|\times|\calS|}$ be a state transition probability matrix, and their components are:
\begin{flalign}
\mathbf{P}_{\pi}[s,s'] =\sum_{a\in\mathcal{A}}\pi(a|s)\mathbb{P}(s'|s,a)=:\Pro_{\pi}(s^{'}|s),
\end{flalign}
which denotes one-step state transformation probability from $s$ to $s^{'}$ by executing $\pi$. To better understand the one-step state transition under a policy $\pi$, we illustrate it in the next Figure \ref{fig:single-step-pro}.
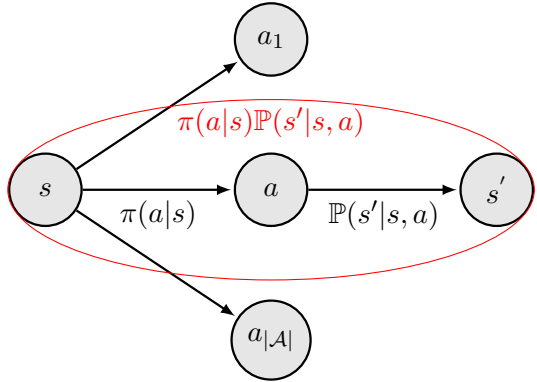
\begin{figure}[h]
\begin{center}
\begin{tikzpicture}[font=\sffamily]

        \tikzset{node style/.style={state, 
                                    minimum width=0.1cm,
                                    line width=0.3mm,
                                    fill=gray!20!white,scale=1}}

        \node[node style] at (0, 0)     (s)     {$s$};
        \node[node style] at (3, 2)  (a1)     {$a_1$};
        \node[node style] at (3, 0)    (a)     {$a$};
        \node[node style] at (3, -2) (al)     {$a_{|\mathcal{A}|}$};
          \node[node style] at (6, 0)     (ss)     {$s^{'}$};
        \draw[every loop,
              auto=right,
              line width=0.3mm,
              >=latex,
              draw=black,
              fill=black]
            (s)      edge  node {$$}  (a1)
              (s)     edge     node {$$} (al)
               (s)     edge     node {${\pi}(a|s)$} (a)
               (a)     edge     node {$\mathbb{P}(s'|s,a)$} (ss);
            \draw[red] ($(a)+(0,0)$) ellipse (3.5cm and 1.2cm)node[yshift=0.9cm]{$\pi(a|s)\mathbb{P}(s'|s,a)$};
\end{tikzpicture}
\end{center}
\caption{Backup diagrams for $\Pro_{\pi}(s^{'}|s)=\sum_{a\in\mathcal{A}}\pi(a|s)\mathbb{P}(s'|s,a)$.}
\label{fig:single-step-pro}
\end{figure}

\subsection{Multi-Step State Transition Probability Matrix}

We are interested in the state distribution induced by a policy.
Recall the following visitation sequence $\tau=\{s_{t}, a_{t}, r_{t+1}\}_{t\ge0}$ induced by ${\pi}$,
we use $\mathbb{P}_{\pi}(s_t=s|s_0)$ to denote the probability of visiting $s$ after $t$
time steps from the initial state $s_0$ by executing $\pi$.
Particularly, we notice if $t=0$, $s_t\ne s_0$, then $\mathbb{P}_{\pi}(s_t=s|s_0)=0$, i.e.,
\begin{flalign}
\label{special-inititial-pro}
\mathbb{P}_{\pi}(s_t=s|s_0)=0,~~ t=0~\text{and}~s\ne s_0.
\end{flalign}

In this lecture, to simplify the expressions, we also use $\mathbb{P}^{(t)}_{{\pi}}$ to denote the notation $\mathbb{P}_{{\pi}}(s_t=s^{'}|s_0=s)$, where $s_0\sim\rho_{0}(\cdot)$, i.e.,
\[
\mathbb{P}^{(t)}_{{\pi}}(s^{'}|s)=:\mathbb{P}_{{\pi}}(s_t=s^{'}|s_0=s).
\]
Furthermore, we use $\bP^{(t)}_{\pi}$ to denote the  $t$-step transition matrix collects all the probability of transition after $t$
time steps by executing ${\pi}$, i.e.,
\[
\bP^{(t)}_{\pi}[s,s^{'}]=\mathbb{P}^{(t)}_{{\pi}}(s^{'}|s).
\]

In order to express the above stochastic process more vividly, we introduce a chain (induced by $\pi$) as follows,
\begin{flalign}
\label{chain-tau-pi}
s_{0}\overset{a_0\sim{\pi}(\cdot|s_0)}{\xrightarrow{\hspace*{1cm}}} \{r_1,s_1\}
\overset{a_1\sim{\pi}(\cdot|s_1)}{\xrightarrow{\hspace*{1cm}}} \{r_2,s_2\}
\overset{a_2\sim{\pi}(\cdot|s_2)}{\xrightarrow{\hspace*{1cm}}}\{r_3,s_3\}\cdots.
\end{flalign}
Particularly, for the chain (\ref{chain-tau-pi}) starting from the initial state $s_0$, the following equity holds
\begin{flalign}
\label{pro-pi-t-step}
\mathbb{P}_{{\pi}}(s_t=s|s_0)=&\sum_{s^{'}\in\mathcal{S}}\mathbb{P}_{{\pi}}(s_t=s|s_{t-1}=s^{'})\mathbb{P}_{{\pi}}(s_{t-1}=s^{'}|s_0),
\end{flalign}
we will show it later.
Due to the Markov property, we know
$
\mathbb{P}_{{\pi}}(s_t=s|s_{t-1}=s^{'})=\mathbb{P}_{{\pi}}(s|s^{'}),
$
then, we rewrite (\ref{pro-pi-t-step}) as follows
\begin{flalign}
\label{pro-pi-t-step-01}
\mathbb{P}_{{\pi}}(s_t=s|s_0)=&\sum_{s^{'}\in\mathcal{S}}\mathbb{P}_{{\pi}}(s|s^{'})\mathbb{P}_{{\pi}}(s_{t-1}=s^{'}|s_0),
\end{flalign}
which can be rewritten as the following concise formulation,
\begin{flalign}
\label{pro-pi-t-step-02}
\mathbb{P}_{{\pi}}^{(t)}(s|s_0)=&\sum_{s^{'}\in\mathcal{S}}\mathbb{P}_{{\pi}}(s|s^{'})\mathbb{P}^{(t-1)}_{{\pi}}(s^{'}|s_0).
\end{flalign}
Eq.(\ref{pro-pi-t-step-02}) is the Chapman-Kolmogorov equation for MDP, we formally present it in the next Theorem \ref{theom:chapman-kolmogorov-eq}, which illustrates the relationship between single-step state transition probability and multi-step state transition probability.

\tcbset{colback=white}
\begin{tcolorbox}
\begin{theorem}[Chapman-Kolmogorov Equation]
\label{theom:chapman-kolmogorov-eq}
Let $\mathbb{P}^{(t)}_{{\pi}}(s^{'}|s)$ be the probability of  transition from state $s$ to state $s^{'}$ after $t$
additional steps by executing a stationary Markovian policy  $\pi$, and  its corresponding $t$-step transition matrix is $\bP^{(t)}_{\pi}$.
Then, 
\begin{flalign}
\label{chapman-kolmogorov-state-to-state}
\mathbb{P}_{{\pi}}^{(t)}(s|s_0)=&\sum_{s^{'}\in\mathcal{S}}\mathbb{P}_{{\pi}}(s|s^{'})\mathbb{P}^{(t-1)}_{{\pi}}(s^{'}|s_0).
\end{flalign}
Furthermore, we know
\begin{flalign}
\label{chapman-kolmogorov-matrix}
\bP_{{\pi}}^{(t)}=\bP_{{\pi}}^{t}.
\end{flalign}
\end{theorem}
\end{tcolorbox}
\begin{proof} We only need to show the result (\ref{pro-pi-t-step}), we give a simple derivation of (\ref{pro-pi-t-step}) follows \cite{weng2018PG},
\begin{itemize}
\item For the case $t=0$, \[\mathbb{P}_{{\pi}}(s_t=s|s_0)=\mathbb{P}_{{\pi}}(s_0=s|s_0)=1,\] which is a trivial fact.
\item For the case $t=1$, we know
\[\mathbb{P}_{{\pi}}(s_1=s|s_0)=\sum_{a\in\mathcal{A}}{\pi}(a|s_0)\Pro(s|s_0,a)=:\E_{a\sim\pi(\cdot|s_0)}[\Pro(s|s_0,a)],\] which is reduced to single state transition probability by executing ${\pi}$, and it is same with the result of (\ref{pro-pi-t-step}). In fact, since the chain (\ref{chain-tau-pi}) starts from the initial state $s_0$, then we have
\begin{flalign}
\nonumber
&\sum_{s^{'}\in\mathcal{S}}\mathbb{P}_{{\pi}}(s_1=s|s_0=s^{'})\mathbb{P}_{{\pi}}(s_0=s^{'}|s_0)\\
\nonumber
=&\sum_{s^{'}\in\mathcal{S}-\{s_0\}}\mathbb{P}_{{\pi}}(s_1=s|s_0=s^{'})\underbrace{\mathbb{P}_{{\pi}}(s_{t-1}=s^{'}|s_0)}_{=0,~\text{if}~t=1;\tau~\text{starts~from}~s_0}+\mathbb{P}_{{\pi}}(s_1=s|s_0)\underbrace{\mathbb{P}_{{\pi}}(s_{t-1}=s_0|s_0)}_{=1;~\text{if}~t=1}.
\end{flalign}
\item For the general case time $t$,
we can first travel from $s_0$ to a middle point $s^{'}$ (any state can be a middle point), after $t-1$ steps, and then go to the final state $s$ during the last step. In this way, we are able to update the visitation probability recursively as (\ref{pro-pi-t-step}).
\end{itemize}
Eq.(\ref{chapman-kolmogorov-matrix}) is a matrix version of Eq.(\ref{chapman-kolmogorov-state-to-state}).
\end{proof}

\subsection{Discounted State Distribution}
Let $d_{{\pi}}^{s_0}(s)$ denote the normalized discounted weighting of the future state $s$ encountered starting at $s_0$ by executing ${\pi}$,
\begin{flalign}
\label{state-distribution-01}
d_{{\pi}}^{s_0}(s)=(1-\gamma)\sum_{t=0}^{\infty}\gamma^{t}\mathbb{P}_{{\pi}}(s_t=s|s_0).
\end{flalign}
Furthermore, since $s_0\sim\rho_{0}(\cdot)$, we define
\begin{flalign}
\label{state-distribution-02}
d_{{\pi}}^{\rho_0}(s)=\mathbb{E}_{s_0\sim\rho_{0}(\cdot)}[d_{{\pi}}^{s_0}(s)]
=\sum_{s_0\in\calS}\rho_{0}(s_0)d^{s_0}_{{\pi}}(s)
=\int_{s_0\in\calS}\rho_{0}(s_0)d^{s_0}_{{\pi}}(s)\text{d}s_0
\end{flalign}
as the discounted state visitation distribution over the initial distribution $\rho_0(\cdot)$.

We use $\bd_{{\pi}}^{\rho_0}\in\R^{|\calS|}$ to collect all the normalized discounted state distributions, and its components are:
\[
\bd_{{\pi}}^{\rho_0}[s]=d_{{\pi}}^{\rho_0}(s),~~s\in\calS.
\]

Recall $\bP_{\pi}\in\R^{|\calS|\times|\calS|}$ denotes the one-step state transition matrix by executing $\pi$, and
we use $\bm{\rho}_0\in\R^{|\calS|}$ denotes the initial state distribution vector, and their components are:
\[
\bP_{\pi}[s,s'] =\sum_{a\in\mathcal{A}}\pi(a|s)\mathbb{P}(s'|s,a),~~\bm{\rho}_{0}[s]=\rho_{0}(s).
\]
Then, we rewrite $\bd_{{\pi}}^{\rho_0}$ as a matrix version as follows,
\begin{flalign}
\label{normalized-discounted-state-distributions-mat}
\bd_{{\pi}}^{\rho_0}=(1-\gamma)\sum_{t=0}^{\infty}(\gamma\bP_{\pi})^{t}\bm{\rho}_{0}=(1-\gamma)(\bI-\gamma\bP_{\pi})^{-1}\bm{\rho}_{0}.
\end{flalign}

\subsection{Reward}

\begin{figure}[t]
\begin{center}
\begin{tikzpicture}[font=\sffamily]

        \tikzset{node style/.style={state, 
                                    minimum width=0.1cm,
                                    line width=0.3mm,
                                    fill=gray!20!white,scale=1}}

        \node[node style] at (0, 0)     (s)     {$s$};
        \node[node style] at (4, 1.8)  (a1)     {$a_1$};
        \node[node style] at (4, 0)    (a)     {$a$};
        \node[node style] at (4, -1.8) (al)     {$a_{|\mathcal{A}|}$};
         \node[node style] at (8, 1.8)     (ss1)     {$s_1$};
          \node[node style] at (8, 0)     (ss)     {$s^{'}$};
            \node[node style] at (8, -1.8)     (ssl)     {$s_{|\mathcal{S}|}$};
        \draw[every loop,
              auto=right,
              line width=0.3mm,
              >=latex,
              draw=black,
              fill=black]
            (s)      edge  node {$$}  (a1)
              (s)     edge     node {$$} (al)
               (s)     edge     node {$$} (a)
               (a)     edge     node {$$} (ss1)
               (a)     edge    node {$$} (ssl)
               (a)     edge     node {$r(s'|s,a)$} (ss);
            \draw[blue] ($(a)+(0,0)$) ellipse (4.8cm and 1.2cm)node[yshift=0.9cm]{$\pi(a|s)\mathbb{P}(s'|s,a)$};
\end{tikzpicture}
\end{center}
\caption{Backup diagrams for $R_{\pi}(s) =\sum_{a\in\mathcal{A}}\sum_{s^{'}\in\mathcal{S}}\pi(a|s)\Pro(s^{'}|s,a)r(s'|s,a)$.}
\label{fig:single-step-reward}
\end{figure}
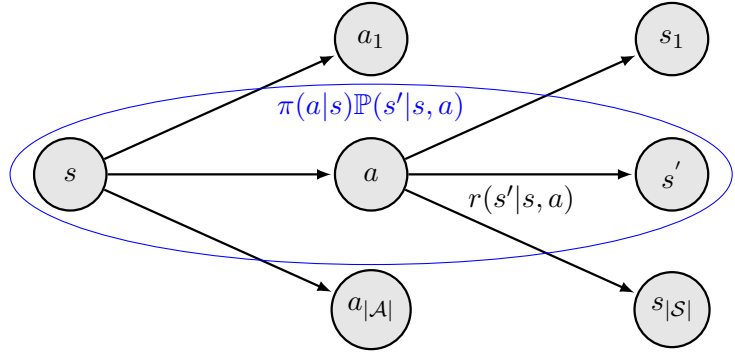

It is noteworthy that
if the reward $r_{t+1}$ depends on the state of the environment at the next state, we use $r(s_{t+1}|s_t,a_t)$ to replace $r_{t+1}$ to denote a real value that the decision-maker receives at time $t$ when the system is at state $s_t$, action $a_t$ is played and the system transforms to the next state $s_{t+1}$.
Then, the expected reward at time $t$ can be evaluated as follows,
\begin{flalign}
\label{reward-fun-depent}
\mathbb{E}[r_{t+1}]=R(s_t,a_t)=\sum_{s_{t+1}\in\cal S}\mathbb{P}(s_{t+1}|s_t,a_t)r(s_{t+1}|s_t,a_t).
\end{flalign}
Under most notions of optimality, all of the information necessary to make a decision at time $t$ is summarized in $r_{t+1}$; however, under some criteria, we must use $r(s_{t+1}|s_t,a_t)$ instead of $r_{t+1}$.

Furthermore, due to the Markov property in the MDP, for each $(s,a)\in\calS\times\calA$, we rewrite (\ref{reward-fun-depent}) as follows:
\begin{flalign}
\label{eq:gen-reward-fun-depent}
R(s,a)=\sum_{s^{'}\in\cal S}\mathbb{P}(s^{'}|s,a)r(s^{'}|s,a).
\end{flalign}

Let $\mathbf{r}_{\pi}\in\R^{|\calS|}$ be the expected reward according to $\pi$, i.e., their components are: $\forall s\in\calS$,
\begin{flalign}
\label{reward-wrt-policy}
\mathbf{r}_{\pi}[s] =\sum_{a\in\mathcal{A}}\sum_{s^{'}\in\mathcal{S}}\pi(a|s)\Pro(s^{'}|s,a)r(s'|s,a)=\sum_{a\in\mathcal{A}}\pi(a|s)R(s,a)=:R_{\pi}(s).
\end{flalign}

Starting from state $s$, the root node at the left, the agent could take any of some set of actions (e.g., $a$), then the environment could respond with one of several next state $s^{'}$, then we obtain the reward $r(s^{'}|s,a)$.
Figure \ref{fig:single-step-reward} has shown the reward after the state transformation from state $s$ to $s^{'}$ by executing $\pi$, which also provides an insight for one-step state transformation probability $R_{\pi}(s)$.

\subsection{Value Function}

The \emph{state value function} of $\pi$ is defined as 
\begin{flalign}V_{\pi}(s) = \mathbb{E}_{\pi}\left[\sum_{t=0}^{\infty}\gamma^{t}r_{t+1}\Big|s_{0} = s\right],
\end{flalign}
where $\mathbb{E}_{\pi}[\cdot|\cdot]$ denotes a conditional expectation on actions which are selected by $\pi$.
Its \emph{state-action value function} is 
\begin{flalign}
Q_{\pi}(s,a) = \mathbb{E}_{\pi}\left[\sum_{t=0}^{\infty}\gamma^{t}r_{t+1}\Big|s_{0} = s,a_{0}=a\right],
\end{flalign} 
and advantage function is 
\begin{flalign}
A_{\pi}(s,a)=Q_{\pi}(s,a) -V_{\pi}(s).
\end{flalign}

\subsection{Bellman Equation}

Bellman equation illustrates the relationship between the states' values and actions, which plays a central role in MDP theory and reinforcement learning.
\tcbset{colback=white}
\begin{tcolorbox}
\begin{theorem}[Bellman Equation]
The state value function $V_{\pi}(s)$ and state-action value function $Q_{\pi}(s,a)$ satisfy the following equation:
\begin{flalign}
V_{\pi}(s)= &R_{\pi}(s)+\gamma \sum_{s^{'}\in\calS}\Pro_{\pi}(s^{'}|s)V_{\pi}(s^{'}),\\
Q_{\pi}(s,a)=&R(s,a)+\gamma\sum_{s^{'}\in\calS}\sum_{a^{'}\in\calA}\Pro(s^{'}|s,a)\pi(a^{'}|s^{'})Q_{\pi}(s^{'},a^{'}).
\end{flalign}
\end{theorem}
\end{tcolorbox}

\begin{proof}
First, we notice 
\begin{flalign}
\nonumber
G_{t}&=:r_{t+1}+\gamma r_{t+2}+\gamma^{2} r_{t+3}+\cdots\\
\nonumber
&=r_{t+1}+\gamma G_{t+1}.
\end{flalign}
Then we rewrite the state value function as follows
\begin{flalign}
\nonumber
V_{\pi}(s) =& \mathbb{E}_{\pi}\left[\sum_{t=0}^{\infty}\gamma^{t}r_{t+1}\Big|s_{0} = s\right]\\
\nonumber
=&\mathbb{E}_{\pi}\left[r_{1}+\gamma G_{1}|s_{0} = s\right].
\end{flalign}

For the first term, we know
\begin{flalign}
\label{temp-02}
\E_{\pi}[r_{1}|s_{0} = s]=\E_{a\sim\pi(\cdot|s),s^{'}\sim\Pro(\cdot|s,a)}[r(s^{'}|s,a)]=\sum_{a\in\calA}\sum_{s^{'}\in\calS}\pi(a|s)\Pro(s^{'}|s,a) r(s^{'}|s,a).
\end{flalign}

For the second term, we know
\begingroup
\allowdisplaybreaks
\begin{flalign}
\nonumber
\E_{\pi}[G_{1}|s_{0} = s]
\nonumber
=&\sum_{s^{'}\in\calS}\Pro_{\pi}(s^{'}|s)\E_{\pi}\left[G_{1}|s_{0} = s,s_{1}=s^{'}\right]\\
\label{temp-01}
=&\sum_{s^{'}\in\calS}\Pro_{\pi}(s^{'}|s)\E_{\pi}\left[G_{1}|s_{1}=s^{'}\right]\\
\nonumber
=&\sum_{s^{'}\in\calS}\Pro_{\pi}(s^{'}|s)V_{\pi}(s^{'})\\
=&\sum_{a\in\calA}\sum_{s^{'}\in\calS}\pi(a|s)\Pro(s^{'}|s,a) V_{\pi}(s^{'}),
\end{flalign}
\endgroup
where Eq.(\ref{temp-01}) holds due to the conditional independence as follows,
\[
\E_{\pi}\left[G_{1}|s_{1}=s^{'}\right]=\E_{\pi}\left[G_{1}|s_{1}=s^{'},s_{0}=s\right].
\]
Such conditional independence property is due to the memoryless Markov property that the future behavior totally depends on the current state.

Then, combining (\ref{temp-02}) and (\ref{temp-01}), we obtain the \emph{Bellman equaiton} as follows,
\begin{flalign}
V_{\pi}(s)=&\underbrace{\sum_{a\in\calA}\sum_{s^{'}\in\calS}\pi(a|s)\Pro(s^{'}|s,a) r(s^{'}|s,a)}_{\text{mean~of~current~rewards}}+\underbrace{\gamma\sum_{a\in\calA}\sum_{s^{'}\in\calS}\pi(a|s)\Pro(s^{'}|s,a) V_{\pi}(s^{'})}_{\text{mean~of~future~rewards}}\\
\overset{(\ref{reward-wrt-policy})}=& R_{\pi}(s)+\gamma \sum_{s^{'}\in\calS}\Pro_{\pi}(s^{'}|s)V_{\pi}(s^{'}).
\end{flalign}

Similarly, we know the state-action value function version of the Bellman equation.
\end{proof}
Finally, we use $\bv_{\pi}\in\R^{|\calS|}$ to collect all the state value functions, and each entry of $\bv_{\pi}$ is defined as \[\bv_{\pi}[s]=V_{{\pi}}(s),\] then we rewrite the Bellman equation as the following matrix version:
\begin{flalign}
\label{eq:bellman-v-mat}
\bv_{\pi}=\br_{\pi}+\gamma\bP_{\pi}\bv_{\pi}=(\bI-\gamma\bP_{\pi})^{-1}\br_{\pi}.
\end{flalign}

\subsection{Objective of Reinforcement Learning}

Recall $\tau=\{s_{t}, a_{t}, r_{t+1}\}_{t\ge0}\sim{\pi}$, according to $\tau$,
we define the expected return $J(\pi |s_0)$ by
\begin{flalign}
\label{Eq:J-theta}
  J(\pi|s_0)=&\mathbb{E}_{\tau\sim{\pi}}[R(\tau)]=V_{\pi}(s_0),
\end{flalign}
where $R(\tau)=\sum_{t\ge0}\gamma^{t}r_{t+1}$, and the notation $J(\pi|s_0)$ is ``conditional'' on $s_0$ is to emphasize the trajectory $\tau$ starting from $s_0$.
Let 
\begin{flalign}
\label{objective-mdp}
J({\pi})=\mathbb{E}_{s_0\sim\rho_{0}(\cdot)}[J(\pi|s_0)]=\mathbb{E}_{s_0\sim\rho_{0}(\cdot)}[V_{\pi}(s_0)]=\sum_{s_0\in\calS}\rho_{0}(s_0)V_{\pi}(s_0).
\end{flalign}
The goal of reinforcement learning is to search the optimal policy $\pi_{\star}$ satisfies 
\begin{flalign}
\pi_{\star}=\arg\max_{\pi}J(\pi).
\end{flalign}

To see the objective (\ref{objective-mdp}) clearly, we use rewrite $J(\pi)$ with respect to $\bd_{{\pi}}^{\rho_0}$.

\tcbset{colback=white}
\begin{tcolorbox}
\begin{theorem} 
\label{them:obj-01}
The objective $J(\pi)$ shares the following versions
\begin{flalign}
\nonumber
J({\pi})=&\sum_{s_0\in\calS}\rho_{0}(s_0)V_{\pi}(s_0)\\
=&
\dfrac{1}{1-\gamma}\sum_{s_0\in\calS}\rho_{0}(s_0)\sum_{s\in\calS}d^{s_0}_{\pi}(s) R_{{\pi}}(s)\\
=&\dfrac{1}{1-\gamma}\mathbb{E}_{s\sim d_{\pi}^{\rho_0}(\cdot),a\sim\pi(\cdot|s),s^{'}\sim\Pro(\cdot|s,a)}\left[r(s^{'}|s,a)\right].
\end{flalign}
Furthermore, the matrix version is
\begin{flalign}
J(\pi)=\bm{\rho}_{0}^{\top}\bv_{\pi}=\bm{\rho}_{0}^{\top}(\bI-\gamma\bP_{\pi})^{-1}\br_{\pi}=\dfrac{1}{1-\gamma}\left\langle\bd_{\pi}^{\rho_0},\br_{\pi}\right\rangle.
\end{flalign}
\end{theorem}
\end{tcolorbox}

\begin{proof}
Recall the Bellman equation, we obtain
\begin{flalign}
\label{bellman-eq-1}
V_{{\pi}}(s_0)
=&\sum_{a\in\mathcal{A}}{\pi}(a|s_0)R(s_0,a)+\gamma\sum_{s^{'}\in\mathcal{S}}\mathbb{P}_{{\pi}}(s_1=s^{'}|s_0)V_{{\pi}}(s^{'}),
\end{flalign}
and we unroll the expression of (\ref{bellman-eq-1}) repeatedly, then we have
\begingroup
\allowdisplaybreaks
\begin{flalign}
\nonumber
V_{{\pi}}(s_0)
=&R_{{\pi}}(s_0)+\gamma\sum_{s^{'}\in\mathcal{S}}\mathbb{P}_{{\pi}}(s_1=s^{'}|s_0)
\underbrace{\left(
R_{\pi}(s^{'})
+\gamma\sum_{s^{''}\in\mathcal{S}}\mathbb{P}_{{\pi}}(s_2=s^{''}|s_1=s^{'})V_{{\pi}}(s^{''})
\right)}_{=V_{{\pi}}(s^{'})}\\
\nonumber
=&R_{{\pi}}(s_0)+\gamma\sum_{s^{'}\in\mathcal{S}}\mathbb{P}_{{\pi}}(s_1=s^{'}|s_0)R_{\pi}(s^{'})\\
\nonumber
&~~~~~~~~~~~~~~+\gamma^2\sum_{s^{''}\in\mathcal{S}}\underbrace{
\left(
\sum_{s^{'}\in\mathcal{S}}\mathbb{P}_{{\pi}}(s_1=s^{'}|s_0)
\mathbb{P}_{{\pi}}(s_2=s^{''}|s_1=s^{'})
\right)
}_{=\mathbb{P}_{{\pi}}(s_2=s^{''}|s_0)}V_{{\pi}}(s^{''})\\
\nonumber
=&R_{{\pi}}(s_0)+\gamma\sum_{s\in\mathcal{S}}\mathbb{P}_{{\pi}}(s_1=s|s_0)R_{{\pi}}(s)
+\gamma^2\sum_{s\in\mathcal{S}}\mathbb{P}_{{\pi}}(s_2=s|s_0)V_{{\pi}}(s)\\
\nonumber
=&R_{{\pi}}(s_0)+\gamma\sum_{s\in\mathcal{S}}\mathbb{P}_{{\pi}}(s_1=s|s_0)R_{{\pi}}(s)\\
\nonumber
&~~~~~~~~~~~~~~+\gamma^2\sum_{s\in\mathcal{S}}\mathbb{P}_{{\pi}}(s_2=s|s_0)
\left(
R_{{\pi}}(s)+\gamma\sum_{s^{'}\in\mathcal{S}}\mathbb{P}_{{\pi}}(s_3=s^{'}|s_2=s)V_{{\pi}}(s^{'})
\right)\\
\nonumber
=&R_{{\pi}}(s_0)+\gamma\sum_{s\in\mathcal{S}}\mathbb{P}_{{\pi}}(s_1=s|s_0)R^{{\pi}}(s)+\gamma^2\sum_{s\in\mathcal{S}}\mathbb{P}_{{\pi}}(s_2=s|s_0)R_{{\pi}}(s)\\
\nonumber
&~~~~~~~~~~~~~~+\gamma^3\sum_{s^{'}\in\mathcal{S}}\underbrace{
\left(
\sum_{s\in\mathcal{S}}\mathbb{P}_{{\pi}}(s_2=s|s_0)
\mathbb{P}_{{\pi}}(s_3=s^{'}|s_2=s)
\right)
}_{=\mathbb{P}_{{\pi}}(s_3=s^{'}|s_0)}V_{{\pi}}(s^{'})\\
\nonumber
=&R_{{\pi}}(s_0)+\gamma\sum_{s\in\mathcal{S}}\mathbb{P}_{{\pi}}(s_1=s|s_0)R^{{\pi}}(s)+\gamma^2\sum_{s\in\mathcal{S}}\mathbb{P}_{{\pi}}(s_2=s|s_0)R^{{\pi}}(s)\\
\nonumber
&~~~~~~~~~~~~~~+\gamma^3\sum_{s\in\mathcal{S}}\mathbb{P}_{{\pi}}(s_3=s|s_0)V_{{\pi}}(s)\\
\nonumber
=&\cdots
\\
\label{re-bellman-eq-01}
=&\sum_{s\in\calS}\sum_{t=0}^{\infty}\gamma^{t}\mathbb{P}_{{\pi}}(s_t=s|s_0)R_{{\pi}}(s)
\overset{(\ref{state-distribution-01})}=\dfrac{1}{1-\gamma}\sum_{s\in\calS}d^{s_0}_{\pi}(s) R_{{\pi}}(s)
.
\end{flalign}
According to (\ref{objective-mdp}) and (\ref{re-bellman-eq-01}), we have
\begin{flalign}
\nonumber
J({\pi})=&\sum_{s_0\in\calS}\rho_{0}(s_0)V_{\pi}(s_0)\overset{(\ref{re-bellman-eq-01})}=
\dfrac{1}{1-\gamma}\sum_{s_0\in\calS}\rho_{0}(s_0)\sum_{s\in\calS}d^{s_0}_{\pi}(s) R_{{\pi}}(s)\\
\nonumber
=&\dfrac{1}{1-\gamma}\sum_{s\in\calS}\underbrace{\left(\sum_{s_0\in\calS}\rho_{0}(s_0)d^{s_0}_{\pi}(s)\right)}_{=d^{\rho_0}_{\pi}(s)} R_{{\pi}}(s)\\
\nonumber
\overset{(\ref{reward-fun-depent})}=&\dfrac{1}{1-\gamma}\sum_{s\in\calS}d^{\rho_0}_{\pi}(s)\sum_{a\in\mathcal{A}}{\pi}(a|s)\sum_{s^{'}\in\calS}\Pro(s^{'}|s,a)r(s^{'}|s,a)\\
\label{re-bellman-eq-02}
=&\dfrac{1}{1-\gamma}\mathbb{E}_{s\sim d_{\pi}^{\rho_0}(\cdot),a\sim\pi(\cdot|s),s^{'}\sim\Pro(\cdot|s,a)}\left[r(s^{'}|s,a)\right].
\end{flalign}
\endgroup

We also have a simple way to achieve the above result,
\begin{flalign}
\label{objec-01}
J(\pi)=\bm{\rho}_{0}^{\top}\bv_{\pi}\overset{(\ref{eq:bellman-v-mat})}=\bm{\rho}_{0}^{\top}(\bI-\gamma\bP_{\pi})^{-1}\br_{\pi}
\overset{(\ref{normalized-discounted-state-distributions-mat})}=\dfrac{1}{1-\gamma}\left\langle\bd_{\pi}^{\rho_0},\br_{\pi}\right\rangle.
\end{flalign}
This concludes the proof of Theorem \ref{them:obj-01}.
\end{proof}

\section{$\lambda$-Return Version of Objective}

In this section, we provide a $\lambda$-return version of the objective, which unifies the version from (\ref{objec-01}).
Before we present our main result, we need some basic results with respect to $\lambda$-return.

\subsection{Bellman Operator}

Let $\mathcal{B}_{{\pi}}$ be the \emph{Bellman operator}:
 \begin{flalign}
 \label{bellman-op}
\mathcal{B}_{{\pi}}:  \mathbb{R}^{|\mathcal{S}|}&\rightarrow \mathbb{R}^{|\mathcal{S}|},\\
 v&\mapsto \mathbf{r}_{\pi}+\gamma \mathbf{P}_{\pi}v,
 \end{flalign}
 where
$\mathbf{r}_{\pi}\in\R^{|\calS|}$ defined in (\ref{reward-wrt-policy}).

Let $\bv_{\pi}\in\R^{|\calS|}$ be a vector that collects all the state value functions, and its components are:
\[
\bv_{\pi}[s]=V_{\pi}(s),~~s\in\calS.
\]
Then, according to Bellman operator (\ref{bellman-op}), we rewrite Bellman equation (\ref{eq:bellman-v-mat}) as the following matrix version:
\begin{flalign}
\label{bellman-eq-matrix}
\mathcal{B}_{{\pi}}\bv_{\pi}=\bv_{\pi}.
\end{flalign}

\subsection{$\lambda$-Bellman Operator}

Furthermore, we define $\lambda$-\emph{Bellman operator} $\mathcal{B}^{\lambda}_{{\pi}}$ as follows,
\[
\mathcal{B}^{\lambda}_{{\pi}}=(1-\lambda)\sum_{t=0}^{\infty}\lambda^{t} (\mathcal{B}_{{\pi}})^{{t}+1},
\]
which implies
 \begin{flalign}
 \label{bellman-op-lam}
\mathcal{B}^{\lambda}_{{\pi}}: \mathbb{R}^{|\mathcal{S}|}\rightarrow& \mathbb{R}^{|\mathcal{S}|},\\
 v\mapsto &\mathbf{r}^{(\lambda)}_{\pi}+\tilde{\gamma}\mathbf{P}^{(\lambda)}_{\pi}v,
 \end{flalign}
 where 
 \begin{flalign}
\label{def:matrix-p-lam-return}
 \mathbf{P}^{(\lambda)}_{\pi}=(1-\gamma\lambda)\sum_{{t}=0}^{\infty}(\gamma\lambda)^{{t}}\bP^{{t}+1}_{{\pi}},~~ \mathbf{r}^{(\lambda)}_{\pi}=\sum_{{t}=0}^{\infty}(\gamma\lambda\bP_{{\pi}})^{{t}}\mathbf{r}_{\pi},~~\tilde{\gamma}=\dfrac{\gamma(1-\lambda)}{1-\gamma\lambda}.
 \end{flalign}

  \begin{remark}[$\lambda$-Return Version of Bellman Equation]
 According to Bellman equation (\ref{bellman-eq-matrix}), $\bv_{\pi}$ is fixed point of $\lambda$-operator $\mathcal{B}^{\lambda}_{{\pi}}$, i.e.,
 \begin{flalign}
 \label{bellman-eq-return}
 \bv_{\pi}=\mathbf{r}^{(\lambda)}_{\pi}+{\tilde{\gamma}} \mathbf{P}^{(\lambda)}_{\pi}\bv_{\pi}.
 \end{flalign}
 Recall $\tau=\{s_t,a_t,r_{t+1}\}_{t\ge0}\sim{\pi}$, according to (\ref{bellman-eq-return}), the value function of initial state $s_0$ is
 \begin{flalign}
  \nonumber
  V_{{\pi}}(s_0)&= \bv_{\pi}[s_0]=\mathbf{r}^{(\lambda)}_{\pi}[s_0]+\tilde \gamma \mathbf{P}^{(\lambda)}_{\pi}\bv_{\pi}[s_0]\\
  \label{bellman-eq-1-1}
 & =R^{(\lambda)}_{\pi}(s_0)+{\tilde{\gamma}}\sum_{s^{'}\in\calS} \Pro_{{\pi}}^{(\lambda)}(s_1=s^{'}|s_0)V_{{\pi}}(s^{'}).
 \end{flalign}
 \end{remark}

 \subsection{$\lambda$-Version of Transition Probability Matrix}
 
 Let
 \begin{flalign}
 \label{lam-pro-value}
 \Pro_{{\pi}}^{(\lambda)}(s^{'}|s)=\mathbf{P}^{(\lambda)}_{\pi}[s,s^{'}]=:(1-\gamma\lambda)\sum_{{t}=0}^{\infty}(\gamma\lambda)^{{t}}\left(\bP^{{t}+1}_{{\pi}}[s,s^{'}]\right),
 \end{flalign}
 where $\bP^{{t}+1}_{{\pi}}[s,s^{'}]$ is the $(s,s^{'})$-th component of matrix $\bP^{{t}+1}_{{\pi}}$, which is the probability of visiting $s^{'}$ after $t+1$ time steps from
 the state $s$ by executing ${\pi}$, i.e.,
  \begin{flalign}
 \label{lam-pro-value-01}
 \bP^{{t}+1}_{{\pi}}[s,s^{'}]= \Pro_{{\pi}}(s_{t+1}=s^{'}|s).
\end{flalign}
Thus, we rewrite  $\Pro_{{\pi}}^{(\lambda)}(s^{'}|s)$ (\ref{lam-pro-value}) as follows
 \begin{flalign}
 \label{lam-pro-value-02}
 \Pro_{{\pi}}^{(\lambda)}(s^{'}|s)=(1-\gamma\lambda)\sum_{{t}=0}^{\infty}(\gamma\lambda)^{{t}}\Pro_{{\pi}}(s_{t+1}=s^{'}|s),~~s\in\calS.
 \end{flalign}

 \begin{remark}
Recall the following visitation sequence $\tau=\{s_{t}, a_{t}, r_{t+1}\}_{t\ge0}$ induced by ${\pi}$,
it is similar to the probability $\mathbb{P}_{{{\pi}}}(s_t=s^{'}|s_0)$, we introduce $\mathbb{P}^{(\lambda)}_{{{\pi}}}(s_t=s^{'}|s_0)$
as the probability of  transition from state $s$ to state $s^{'}$after $t$
time steps under the dynamic transformation matrix $ \mathbf{P}^{(\lambda)}_{\pi}$.
Then, the following equity holds
\begin{flalign}
\mathbb{P}_{{{\pi}}}^{(\lambda)}(s_t=s|s_0)=&\sum_{s^{'}\in\mathcal{S}}\mathbb{P}_{{{\pi}}}^{(\lambda)}(s_t=s|s_{t-1}=s^{'})\mathbb{P}_{{{\pi}}}^{(\lambda)}(s_{t-1}=s^{'}|s_0).
\end{flalign}
\end{remark}

 \subsection{$\lambda$-Version of Reward}

 Similarly, let
  \begin{flalign}
  \nonumber
 R^{(\lambda)}_{\pi}(s)=:
  \mathbf{r}^{(\lambda)}_{\pi}[s]=&\sum_{{t}=0}^{\infty}(\gamma\lambda\bP_{{\pi}})^{{t}}\mathbf{r}_{\pi}[s]
  =  \sum_{{t}=0}^{\infty}(\gamma\lambda)^{t}\left(\sum_{s^{'}\in\calS}\Pro_{{\pi}}(s_{t}=s^{'}|s)R_{{\pi}}(s^{'})\right)\\
   \label{lam-pro-value-03}
   =&
     \sum_{{t}=0}^{\infty}\sum_{s^{'}\in\calS}(\gamma\lambda)^{t}\Pro_{{\pi}}(s_{t}=s^{'}|s)R_{{\pi}}(s^{'}).
 \end{flalign}

 \subsection{$\lambda$-Version of Discounted State Distribution}

It is similar to normalized discounted distribution $d_{\pi}^{\rho_0}(s)$, we introduce $\lambda$-return version of discounted state distribution $d_{\pi}^{\lambda}(s)$ as follows: $\forall s\in\calS$,
\begin{flalign}
\label{lambda-dis-state-distribution}
d_{\pi}^{s_0,\lambda}(s)&=(1-\tilde \gamma)\sum_{t=0}^{\infty}\tilde{\gamma}^{t}\mathbb{P}^{(\lambda)}_{\pi}(s_t=s|s_0),\\
d_{\pi}^{\lambda}(s)&=\E_{s_0\sim\rho_{0}(\cdot)}\left[d_{\pi}^{s_0,\lambda}(s)\right],\\
\label{mat-lambda-dis-state-distribution}
\bd_{\pi}^{\lambda}[s]&=d_{\pi}^{\lambda}(s),
\end{flalign}
where $\mathbb{P}^{(\lambda)}_{\pi}(s_t=s|s_0)$ is the $(s_0,s)$-th component of the matrix $\left(\mathbf{P}^{(\lambda)}_{\pi}\right)^{t}$, i.e.,
\[
\mathbb{P}^{(\lambda)}_{\pi}(s_t=s|s_0)=:\left(\mathbf{P}^{(\lambda)}_{\pi}\right)^{t}[s_0,s].
\]
Similarly, $\mathbb{P}^{(\lambda)}_{\pi}(s_t=s^{'}|s)$ is the $(s,s^{'})$-th component of the matrix $\left(\mathbf{P}^{(\lambda)}_{\pi}\right)^{t}$, i.e.,
\[
\mathbb{P}^{(\lambda)}_{\pi}(s_t=s^{'}|s)=:\left(\mathbf{P}^{(\lambda)}_{\pi}\right)^{t}[s,s^{'}].
\]
Finally, we rewrite $\bd_{\pi}^{\lambda}$ as the following matrix version,
\begin{flalign}
\label{matrixversion-lambda-dis-state-distribution}
\bd_{\pi}^{\lambda}=(1-\tilde \gamma)\sum_{t=0}^{\infty}\left(\gamma\bP^{(\lambda)}_{{\pi}}\right)^{t}\bm{\rho}_{0}=(1-\tilde \gamma)\left(\bI-\tilde{\gamma}\bP^{(\lambda)}_{{\pi}}\right)^{-1}\bm{\rho}_{0}.
\end{flalign}

\subsection{$\lambda$-Return Version of Objective}

\tcbset{colback=white}
\begin{tcolorbox}
\begin{theorem}
\label{lem:lam-return-objective}
The objective $J({\pi})$ (\ref{objective-mdp}) can be rewritten as the following version:
\[
J({{\pi}})=\dfrac{1}{1-{\tilde{\gamma}}}\sum_{s\in\calS}d^{\lambda}_{{\pi}}(s)R^{(\lambda)}_{\pi}(s)=
\dfrac{1}{1-{\tilde{\gamma}}}\E_{s\sim d^{\lambda}_{{\pi}}(\cdot)}
\left[R^{(\lambda)}_{\pi}(s)\right].
\]
\end{theorem}
 \end{tcolorbox}

 \begin{proof}
 We unroll the expression of (\ref{bellman-eq-1-1}) repeatedly, then we have
\begingroup
\allowdisplaybreaks
\begin{flalign}
\nonumber
V_{{{\pi}}}(s_0)
=&{R}^{(\lambda)}_{{\pi}}(s_0)+{\tilde{\gamma}}\sum_{s^{'}\in\mathcal{S}}\mathbb{P}_{{{\pi}}}^{(\lambda)}(s_1=s^{'}|s_0)
\underbrace{\left(
{R}^{(\lambda)}_{{\pi}}(s^{'})
+{\tilde{\gamma}}\sum_{s^{''}\in\mathcal{S}}\mathbb{P}_{{{\pi}}}^{(\lambda)}(s_2=s^{''}|s_1=s^{'})V_{{{\pi}}}(s^{''})
\right)}_{=V_{{{\pi}}}(s^{'})}\\
\nonumber
=&{R}^{(\lambda)}_{{\pi}}(s_0)+{\tilde{\gamma}}\sum_{s^{'}\in\mathcal{S}}\mathbb{P}_{{{\pi}}}^{(\lambda)}(s_1=s^{'}|s_0){R}^{(\lambda)}_{{\pi}}(s^{'})\\
\nonumber
&~~~~~~~~~~~~~+{\tilde{\gamma}}^2\sum_{s^{''}\in\mathcal{S}}\underbrace{
\left(
\sum_{s^{'}\in\mathcal{S}}\mathbb{P}_{{{\pi}}}^{(\lambda)}(s_1=s^{'}|s_0)
\mathbb{P}_{{{\pi}}}^{(\lambda)}(s_2=s^{''}|s_1=s^{'})
\right)
}_{\overset{(\ref{pro-pi-t-step})}=:\mathbb{P}^{(\lambda)}_{{{\pi}}}\left(s_2=s^{''}|s_0\right)}V_{{{\pi}}}(s^{''})\\
\nonumber
=&{R}^{(\lambda)}_{{\pi}}(s_0)+{\tilde{\gamma}}\sum_{s\in\mathcal{S}}\mathbb{P}_{{{\pi}}}^{(\lambda)}(s_1=s|s_0){R}^{(\lambda)}_{{\pi}}(s)
+{\tilde{\gamma}}^2\sum_{s\in\mathcal{S}}\mathbb{P}_{{{\pi}}}^{(\lambda)}(s_2=s|s_0)V_{{{\pi}}}(s)\\
\nonumber
=&{R}^{(\lambda)}_{{\pi}}(s_0)+{\tilde{\gamma}}\sum_{s\in\mathcal{S}}\mathbb{P}_{{{\pi}}}^{(\lambda)}(s_1=s|s_0){R}^{(\lambda)}_{{\pi}}(s)\\
\nonumber
&~~~~~~~~~~~~~+{\tilde{\gamma}}^2\sum_{s\in\mathcal{S}}\mathbb{P}_{{{\pi}}}^{(\lambda)}(s_2=s|s_0)
\left(
{R}^{(\lambda)}_{{\pi}}(s)+{\tilde{\gamma}}\sum_{s^{'}\in\mathcal{S}}\mathbb{P}_{{{\pi}}}^{(\lambda)}(s_3=s^{'}|s_2=s)V_{{{\pi}}}(s^{'})
\right)\\
\nonumber
=&{R}^{(\lambda)}_{{\pi}}(s_0)+{\tilde{\gamma}}\sum_{s\in\mathcal{S}}\mathbb{P}_{{{\pi}}}^{(\lambda)}(s_1=s|s_0){R}^{(\lambda)}_{{\pi}}(s)+{\tilde{\gamma}}^2\sum_{s\in\mathcal{S}}\mathbb{P}_{{{\pi}}}^{(\lambda)}(s_2=s|s_0){R}^{(\lambda)}_{{\pi}}(s)\\
\nonumber
&~~~~~~~~~~~~~+{\tilde{\gamma}}^3\sum_{s^{'}\in\mathcal{S}}\underbrace{
\left(
\sum_{s\in\mathcal{S}}\mathbb{P}_{{{\pi}}}^{(\lambda)}(s_2=s|s_0)
\mathbb{P}_{{{\pi}}}^{(\lambda)}(s_3=s^{'}|s_2=s)
\right)
}_{=\mathbb{P}_{{{\pi}}}^{(\lambda)}(s_3=s^{'}|s_0)}V_{{{\pi}}}(s^{'})\\
\nonumber
=&{R}^{(\lambda)}_{}(s_0)+{\tilde{\gamma}}\sum_{s\in\mathcal{S}}\mathbb{P}_{{{\pi}}}^{(\lambda)}(s_1=s|s_0){R}^{(\lambda)}_{{\pi}}(s)+{\tilde{\gamma}}^2\sum_{s\in\mathcal{S}}\mathbb{P}_{{{\pi}}}^{(\lambda)}(s_2=s|s_0){R}^{(\lambda)}_{{\pi}}(s)\\
\nonumber
&~~~~~~~~~~~~~+{\tilde{\gamma}}^3\sum_{s\in\mathcal{S}}\mathbb{P}^{(\lambda)}_{{{\pi}}}(s_3=s|s_0)V_{{{\pi}}}(s)\\
\nonumber
=&\cdots
\\
=&\sum_{s\in\calS}\sum_{t=0}^{\infty}{\tilde{\gamma}}^{t}\mathbb{P}_{{{\pi}}}^{(\lambda)}(s_t=s|s_0){R}^{(\lambda)}_{{\pi}}(s)\\
\label{re-bellman-eq-01-1}
\overset{(\ref{lambda-dis-state-distribution})}=&\dfrac{1}{1-{\tilde{\gamma}}}\sum_{s\in\calS}d^{s_0,\lambda}_{{\pi}}(s) {R}^{(\lambda)}_{{\pi}}(s)
.
\end{flalign}
\endgroup
According to (\ref{objective-mdp}) and (\ref{re-bellman-eq-01-1}), we have
\begin{flalign}
\nonumber
J({{\pi}})=&\sum_{s_0\in\calS}\rho_{0}(s_0)V_{{\pi}}(s_0)\\
\nonumber
\overset{(\ref{re-bellman-eq-01-1})}=&\dfrac{1}{1-{\tilde{\gamma}}}\sum_{s_0\in\calS}\rho_{0}(s_0)\sum_{s\in\calS}d^{s_0,\lambda}_{{\pi}}(s) R^{(\lambda)}_{\pi}(s)\\
\nonumber
=&\dfrac{1}{1-{\tilde{\gamma}}}\sum_{s\in\calS}\underbrace{\left(\sum_{s_0\in\calS}\rho_{0}(s_0)d^{s_0,\lambda}_{{\pi}}(s)\right)}_{=d^{\lambda}_{{\pi}}(s)} R^{(\lambda)}_{\pi}(s)\\
\nonumber
=&\dfrac{1}{1-{\tilde{\gamma}}}\sum_{s\in\calS}d^{\lambda}_{{\pi}}(s)R^{(\lambda)}_{\pi}(s)\\
\label{lam-return-objective}
=&
\dfrac{1}{1-{\tilde{\gamma}}}\E_{s\sim d^{\lambda}_{{\pi}}(\cdot)}
\left[R^{(\lambda)}_{\pi}(s)\right]
.
\end{flalign}
This concludes the proof of Theorem \ref{lem:lam-return-objective}.
\end{proof}

\begin{remark}[Unification]
If $\lambda\rightarrow0$, then Theorem \ref{lem:lam-return-objective} is reduced to Theorem \ref{them:obj-01}.
\end{remark}

\section{A General Version of Objective}

\subsection{Main Result}

\tcbset{colback=white}
\begin{tcolorbox}
\begin{theorem}[\citep{yang2022constrained}]
\label{objective-td-error-version}
For any function $\varphi(\cdot):\calS\rightarrow\R$, for any policy $\policy$, for any trajectory satisfies $\tau=\{s_{t}, a_{t}, r_{t+1}\}_{t\ge0}\sim{\pi}$,
let 
\begin{flalign}
\nonumber
\delta_t^{\varphi}&=r(s_{t+1}|s_t,a_t)+\gamma\varphi(s_{t+1})-\varphi(s_{t}),
\\
\nonumber
\delta^{\varphi}_{\policy,t}(s)&=\E_{s_{t}\sim\Pro_{\policy}(\cdot|s),a_{t}\sim{\policy}(\cdot|s_t),s_{t+1}\sim\Pro(\cdot|s_t,a_t)}\left[\delta_t^{\varphi}\right],
\end{flalign}
then, the objective $J(\policy)$ (\ref{lam-return-objective}) can be rewritten as the following version:
\begin{flalign}
\label{lam-return-phi-objective-prop}
J(\policy)=&\E_{s_0\sim\rho_{0}(\cdot)}[\varphi(s_0)]
+
\dfrac{1}{1-\tilde\gamma}\sum_{s\in\calS}d^{\lambda}_{\policy}(s)
\left(
\sum_{t=0}^{\infty}\gamma^t \lambda^t
\delta^{\varphi}_{\policy,t}(s)
\right)
\\
=&\E_{s_0\sim\rho_{0}(\cdot)}[\varphi(s_0)]
+
\dfrac{1}{1-\tilde\gamma}\E_{s\sim d^{\lambda}_{\policy}(\cdot)}
\left[
\sum_{t=0}^{\infty}\gamma^t \lambda^t
\delta^{\varphi}_{\policy,t}(s)
\right]
.
\end{flalign}
We introduce a vector $\bm{\delta}^{\varphi}_{\policy,t}\in\R^{|\calS|}$ and its components are: for any $s\in\calS$,
\[
\bm{\delta}^{\varphi}_{\policy,t}[s]=:{{\delta}}^{\varphi}_{\policy,t}(s).
\]
Then, we rewrite the objective as the following vector version
\begin{flalign}
\label{lam-return-phi-objective-vec-version}
J(\policy)=\E_{s_0\sim\rho_{0}(\cdot)}[\varphi(s_0)]
+
\dfrac{1}{1-\tilde \gamma}\sum_{t=0}^{\infty}\gamma^t \lambda^t
\langle
\bd_{\pi}^{\lambda},\bm{\delta}^{\varphi}_{\policy,t}
\rangle.
\end{flalign}
\end{theorem}
\end{tcolorbox}

\begin{proof} We show it by the following three steps.

{
\color{blue}
{
\textbf{Step 1: Rewrite the objective $J(\policy)$ in Eq.(\ref{lam-return-objective}).}
}
}

We rewrite the discounted distribution $\bd_{\pi}^{\lambda}$ (\ref{matrixversion-lambda-dis-state-distribution}) as follows,
\begin{flalign}
\label{vector:state-distribution}
\bm{\rho}_{0}-\dfrac{1}{1-{\tilde{\gamma}}}\bd_{\policy}^{\lambda}+\dfrac{{\tilde{\gamma}}}{1-{\tilde{\gamma}}}\bP^{(\lambda)}_{\policy}\bd_{\policy}^{\lambda}=\bm{0}.
\end{flalign}
Let $\varphi(\cdot)$ be a real number function defined on the state space $\calS$, i.e., $\varphi:\calS\rightarrow\R$.
Then we define a vector function $\bm{\phi}(\cdot)\in\R^{|\calS|}$ to collect all the values  $\{\varphi(s)\}_{s\in\calS}$, and its components are 
\[
\bm{\phi}[s]=\varphi(s),~~s\in\calS.
\]
Now, we take the inner product between the vector $\bm{\phi}$ and (\ref{vector:state-distribution}), we have
\begin{flalign}
\nonumber
0&=\left\langle \bm{\rho}_{0}-\dfrac{1}{1-{\tilde{\gamma}}}\bd_{\policy}^{\lambda}+\dfrac{{\tilde{\gamma}}}{1-{\tilde{\gamma}}}\bP^{(\lambda)}_{\policy}\bd_{\policy}^{\lambda},\bm{\phi} \right\rangle
\\
\label{state-distribution-inner-initial-vec}
&=
\langle \bm{\rho}_{0},\bm{\phi}\rangle
-\dfrac{1}{1-{\tilde{\gamma}}}\left\langle \bd_{\policy}^{\lambda},\bm{\phi}\right\rangle
+
\dfrac{{\tilde{\gamma}}}{1-{\tilde{\gamma}}}\left\langle\bP^{(\lambda)}_{\policy}\bd_{\policy}^{\lambda},\bm{\phi}\right\rangle.
\end{flalign}
We express the first term $\langle \bm{\rho}_{0},\bm{\phi}\rangle$ of (\ref{state-distribution-inner-initial-vec}) as follows,
\begin{flalign}
\label{first-term}
\langle \bm{\rho}_{0},\bm{\phi}\rangle=\sum_{s\in\calS}\rho_{0}(s)\varphi(s)=\E_{s\sim\rho_{0}(\cdot)}[\varphi(s)].
\end{flalign}
We express the second term $\langle \bd_{\policy}^{\lambda},\bm{\phi}\rangle$ of (\ref{state-distribution-inner-initial-vec}) as follows,
\begin{flalign}
\label{sec-term}
-\dfrac{1}{1-{\tilde{\gamma}}}\langle \bd_{\policy}^{\lambda},\bm{\phi}\rangle=-\dfrac{1}{1-{\tilde{\gamma}}}\sum_{s\in\calS} d_{\policy}^{\lambda} (s)\varphi(s)
=-\dfrac{1}{1-{\tilde{\gamma}}}\E_{s\sim d_{\policy}^{\lambda} (\cdot)} [\varphi(s)].
\end{flalign}
We express the third term $\langle {\tilde{\gamma}}\bP^{(\lambda)}_{\policy}\bd_{\policy}^{\lambda},\bm{\phi}\rangle$ of (\ref{state-distribution-inner-initial-vec}) as follows,
\begin{flalign}
\dfrac{{\tilde{\gamma}}}{1-{\tilde{\gamma}}}\langle\bP^{(\lambda)}_{\policy}\bd_{\policy}^{\lambda},\bm{\phi}\rangle=&\dfrac{{\tilde{\gamma}}}{1-{\tilde{\gamma}}}\sum_{s^{'}\in\calS}\left(\bP^{(\lambda)}_{\policy}\bd_{\policy}^{\lambda}\right)[s^{'}]\varphi(s^{'})\\
\label{third-term}
=&\dfrac{{\tilde{\gamma}}}{1-{\tilde{\gamma}}}
\sum_{s^{'}\in\calS}
\left(
\sum_{s\in\calS} \Pro_{\policy}^{(\lambda)}(s^{'}|s)d_{\policy}^{\lambda}(s)
\right)
\varphi(s^{'}).
\end{flalign}
According to Theorem \ref{lem:lam-return-objective}, put the results (\ref{lam-return-objective}) and (\ref{state-distribution-inner-initial-vec}) together, we have
\begin{flalign}
\nonumber
J(\policy)\overset{(\ref{lam-return-objective}),(\ref{state-distribution-inner-initial-vec})}=&\dfrac{1}{1-{\tilde{\gamma}}}\sum_{s\in\calS}d^{\lambda}_{\policy}(s)R^{(\lambda)}_{\pi}(s)+\langle \bm{\rho}_{0}-\dfrac{1}{1-{\tilde{\gamma}}}\bd_{\policy}^{\lambda}+\dfrac{{\tilde{\gamma}}}{1-{\tilde{\gamma}}}\bP^{(\lambda)}_{\policy}\bd_{\policy}^{\lambda},\bm{\phi} \rangle\\
\label{lam-return-objective-01}
=&\E_{s_0\sim\rho_{0}(\cdot)}[\varphi(s_0)]
+
\dfrac{1}{1-{\tilde{\gamma}}}\sum_{s\in\calS}d^{\lambda}_{\policy}(s)
\left(
R^{(\lambda)}_{\pi}(s)+{\tilde{\gamma}}
\sum_{s^{'}\in\calS} \Pro_{\policy}^{(\lambda)}(s^{'}|s)\varphi(s^{'})
-\varphi(s)
\right),
\end{flalign}
where the last equation holds since we unfold (\ref{state-distribution-inner-initial-vec}) according to (\ref{first-term})-(\ref{third-term}).

Finally, we should emphasize that the key idea from (\ref{vector:state-distribution}) to (\ref{state-distribution-inner-initial-vec}) is inspired by CPO \citep{achiam2017constrained}.

{
\color{blue}
{
\textbf{Step 2: Rewrite the term $\left(
R^{(\lambda)}_{\pi}(s)+{\tilde{\gamma}}
\sum_{s^{'}\in\calS} \Pro_{\policy}^{(\lambda)}(s^{'}|s)\varphi(s^{'})
-\varphi(s)
\right)$ in Eq.(\ref{lam-return-objective-01}).}
}
}

Then, we unfold the second term of (\ref{lam-return-objective-01}) as follows,
\begingroup
\allowdisplaybreaks
\begin{flalign}
\label{app-04}
&R^{(\lambda)}_{\pi}(s)+{\tilde{\gamma}}
\sum_{s^{'}\in\calS} \Pro_{\policy}^{(\lambda)}(s^{'}|s)\varphi(s^{'})
-\varphi(s)
\\
\nonumber
\overset{(\ref{lam-pro-value-02}),(\ref{lam-pro-value-03})}=&\sum_{{t}=0}^{\infty}({{\gamma}}\lambda\bP_{\policy})^{{t}}\mathbf{r}_{\pi}[s]
+{\tilde{\gamma}}
(1-\gamma\lambda)\sum_{s^{'}\in\calS} \sum_{{t}=0}^{\infty}({{\gamma}}\lambda)^{{t}}\left(\bP^{{t}+1}_{\policy}[s,s^{'}]\right)\varphi(s^{'})
-\varphi(s)\\
\overset{(\ref{def:matrix-p-lam-return})}=&
\sum_{{t}=0}^{\infty}({{\gamma}}\lambda\bP_{\policy})^{{t}}\mathbf{r}_{\pi}[s]
+{{\gamma}}
(1-\lambda)\sum_{s^{'}\in\calS} \sum_{{t}=0}^{\infty}({{\gamma}}\lambda)^{{t}}\Pro_{\policy}(s_{t+1}=s^{'}|s)\varphi(s^{'})
-\varphi(s).
\end{flalign}
\endgroup
Recall the terms $ \mathbf{P}^{(\lambda)}_{\pi},~\mathbf{r}^{(\lambda)}_{\pi}[s]$ defined in (\ref{def:matrix-p-lam-return}), (\ref{lam-pro-value-03}),
\begin{flalign}
\label{app-004}
R^{(\lambda)}_{\pi}(s)+\gamma(1-\lambda)\sum_{s^{'}\in\calS} \Pro_{\policy}^{(\lambda)}(s^{'}|s)\varphi(s^{'})-\varphi(s)
\end{flalign}
We consider the first term $R^{(\lambda)}_{\pi}(s)$ of (\ref{app-04}) as follows,
\begin{flalign}
\label{app-005}
R^{(\lambda)}_{\pi}(s)\overset{(\ref{def:matrix-p-lam-return}),(\ref{lam-pro-value-03})}=
\mathbf{r}^{(\lambda)}_{\pi}[s]=\sum_{{t}=0}^{\infty}(\gamma\lambda)^{t}\bP_{\policy}^{{t}}\mathbf{r}_{\pi}[s]= \sum_{{t}=0}^{\infty}\sum_{s_t\in\calS}(\gamma\lambda)^{t}\Pro_{\policy}(s_{t}|s)R_{\policy}(s_t).
\end{flalign}
We consider the second term $\tilde\gamma\sum_{s\in\calS} \Pro_{\policy}^{(\lambda)}(s^{'}|s)\varphi(s)-\varphi(s)$ of (\ref{app-04}) as follows,
\begingroup
\allowdisplaybreaks
\begin{flalign}
\nonumber
&\tilde\gamma\sum_{s^{'}\in\calS} \Pro_{\policy}^{(\lambda)}(s^{'}|s)\varphi(s^{'})-\varphi(s)\\
\overset{(\ref{lam-pro-value-02})}=&\tilde\gamma
(1-\gamma\lambda)\sum_{s^{'}\in\calS} \sum_{{t}=0}^{\infty}(\gamma\lambda)^{{t}}\Pro_{\policy}(s_{t+1}=s^{'}|s)\varphi(s^{'})
-\varphi(s)\\
\overset{(\ref{def:matrix-p-lam-return})}
=&\gamma
(1-\lambda)\sum_{s^{'}\in\calS} \sum_{{t}=0}^{\infty}(\gamma\lambda)^{{t}}\Pro_{\policy}(s_{t+1}=s^{'}|s)\varphi(s^{'})
-\varphi(s)\\
\nonumber
=&\gamma\sum_{s^{'}\in\calS} \sum_{{t}=0}^{\infty}(\gamma\lambda)^{{t}}\Pro_{\policy}(s_{t+1}=s^{'}|s)\varphi(s^{'})
-\sum_{s^{'}\in\calS} 
\underbrace{\left(\sum_{{t}=0}^{\infty}(\gamma\lambda)^{{t}+1}\Pro_{\policy}(s_{t+1}=s^{'}|s)\varphi(s^{'})\right)}_{=\sum_{{t}=1}^{\infty}(\gamma\lambda)^{{t}}\Pro_{\policy}(s_{t}=s^{'}|s)\varphi(s^{'})}
-\varphi(s)\\
\label{app-001}
=&\gamma\sum_{s^{'}\in\calS} \sum_{{t}=0}^{\infty}(\gamma\lambda)^{{t}}\Pro_{\policy}(s_{t+1}=s^{'}|s)\varphi(s^{'})
-
\underbrace
{
\left(
\sum_{s^{'}\in\calS} \sum_{{t}=1}^{\infty}(\gamma\lambda)^{{t}}\Pro_{\policy}(s_{t}=s^{'}|s)\varphi(s^{'})+\varphi(s)
\right)
}_{=\sum_{s^{'}\in\calS} \sum_{{t}=0}^{\infty}(\gamma\lambda)^{{t}}\Pro_{\policy}(s_{t}=s^{'}|s)\varphi(s^{'})}
\\
\label{app-003}
=&\gamma\sum_{s^{'}\in\calS} \sum_{{t}=0}^{\infty}(\gamma\lambda)^{{t}}\Pro_{\policy}(s_{t+1}=s^{'}|s)\varphi(s^{'})
-\sum_{s_t\in\calS} \sum_{{t}=0}^{\infty}(\gamma\lambda)^{{t}}\Pro_{\policy}(s_{t}|s)\varphi(s),
\end{flalign}
where the equation from Eq.(\ref{app-001}) to Eq.(\ref{app-003}) holds since: according to (\ref{special-inititial-pro}), we use the following identity 
\begin{flalign}
\nonumber
\sum_{s^{'}\in\calS} \Pro_{\policy}(s_{0}=s^{'}|s)\varphi(s^{'})=\varphi(s).
\end{flalign}
\endgroup
Furthermore,  take the result (\ref{app-005}) and (\ref{app-003}) to (\ref{app-004}), we have
\begingroup
\allowdisplaybreaks
\begin{flalign}
\nonumber
&R^{(\lambda)}_{\pi}(s)+\tilde\gamma
\sum_{s^{'}\in\calS} \Pro_{\policy}^{(\lambda)}(s^{'}|s)\varphi(s^{'})-\varphi(s)\\
\label{app-020}
=& \sum_{{t}=0}^{\infty}(\gamma\lambda)^{t}
\left(
\sum_{s_t\in\calS}\Pro_{\policy}(s_{t}|s)
R_{\policy}(s_t)+\gamma\sum_{s^{'}\in\calS}
\underbrace{\Pro_{\policy}(s_{t+1}=s^{'}|s)\varphi(s^{'})}_{\overset{(\ref{pro-pi-t-step-01})}=
\sum_{s_t\in\calS}
\Pro_{\policy}(s_{t+1}=s^{'}|s_t)\Pro_{\policy}(s_t|s)\varphi(s^{'})}
-\sum_{s_{t}\in\calS}\Pro_{\policy}(s_{t}|s)\varphi(s_t)
\right)
\\
\label{app-021}
=& \sum_{{t}=0}^{\infty}(\gamma\lambda)^{t}
\left(
\sum_{s_t\in\calS}\Pro_{\policy}(s_{t}|s)R_{\policy}(s_t)+\gamma\sum_{s_t\in\calS}\Pro_{\policy}(s_{t}|s)\sum_{s_{t+1}\in\calS}\Pro_{\policy}(s_{t+1}|s_{t})\varphi(s_{t+1})
-\sum_{s_t\in\calS}\Pro_{\policy}(s_{t}|s)\varphi(s_t)
\right)
\\
\nonumber
= &\sum_{{t}=0}^{\infty}(\gamma\lambda)^{t}\sum_{s_t\in\calS}
\Pro_{\policy}(s_{t}|s)
\left(
\underbrace{
\sum_{a_t\in\mathcal{A}}{\policy}(a_t|s_t)\sum_{s_{t+1}\in\calS}\Pro(s_{t+1}|s_t,a_t)r(s_{t+1}|s_t,a_t)
}_{=R_{\policy}(s_t)}
\right.
\\
\nonumber
&\left. ~~~~~~~~~~~~~~~~~~~~~~~~~~~~~~~~~~~~~~~~~~~+\gamma\underbrace{\sum_{a_t\in\mathcal{A}}{\policy}(a_t|s_t)\sum_{s_{t+1}\in\calS}\Pro(s_{t+1}|s_t,a_t)}_{=\Pro_{\policy}(s_{t+1}|s_{t})}\varphi(s_{t+1})
-\varphi(s_{t})
\right)\\
\label{app-022}
=& \sum_{{t}=0}^{\infty}(\gamma\lambda)^{t}\sum_{s_t\in\calS}\Pro_{\policy}(s_{t}|s)\sum_{a_t\in\mathcal{A}}{\policy}(a_t|s_t)\sum_{s_{t+1}\in\calS}\Pro(s_{t+1}|s_t,a_t)
\left(r(s_{t+1}|s_t,a_t)+\gamma\varphi(s_{t+1})-\varphi(s_{t})\right)\\
\label{app-023}
=& \sum_{{t}=0}^{\infty}(\gamma\lambda)^{t}\E_{s_{t}\sim\Pro_{\policy}(\cdot|s),a_{t}\sim{\policy}(\cdot|s_t),s_{t+1}\sim\Pro(\cdot|s_t,a_t)}\left[r(s_{t+1}|s_t,a_t)+\gamma\varphi(s_{t+1})-\varphi(s_{t})\right],
\end{flalign}
\endgroup
the equation from Eq.(\ref{app-003}) to Eq.(\ref{app-020}) holds since:
\[
\Pro_{\policy}(s_{t+1}|s)\overset{(\ref{pro-pi-t-step-01})}=
\sum_{s_t\in\calS}
\Pro_{\policy}(s_{t+1}|s_t)\Pro_{\policy}(s_t|s);
\]
the equation from Eq.(\ref{app-020}) to Eq.(\ref{app-021}) holds since we use the Markov property of the definition of MDP: for each time $t\in\N$,
\[\Pro_{\policy}(s_{t+1}=s^{'}|s_t=s)=\Pro_{\policy}(s^{'}|s);\]
the equation (\ref{app-022}) the following identity:
\[
\sum_{a_t\in\mathcal{A}}{\policy}(a_t|s_t)=1,~~~~\sum_{s_{t+1}\in\calS}\Pro(s_{t+1}|s_t,a_t)=1,
\]
then
\[
\varphi(s_{t})=\sum_{a_t\in\mathcal{A}}{\policy}(a_t|s_t)\sum_{s_{t+1}\in\calS}\Pro(s_{t+1}|s_t,a)\varphi(s_{t}).
\]

{
\color{blue}
{
\textbf{Step 3: Put all the results together.}
}
}

Finally, let 
\begin{flalign}
\nonumber
\delta_t^{\varphi}&=r(s_{t+1}|s_t,a_t)+\gamma\varphi(s_{t+1})-\varphi(s_{t}),
\\
\nonumber
\delta^{\varphi}_{\policy,t}(s)&=\E_{s_{t}\sim\Pro_{\policy}(\cdot|s),a_{t}\sim{\policy}(\cdot|s_t),s_{t+1}\sim\Pro(\cdot|s_t,a_t)}\left[\delta_t^{\varphi}\right],
\end{flalign}
combining the results (\ref{lam-return-objective-01}) and (\ref{app-023}), we have
\begin{flalign}
\label{lam-return-phi-objective-prop}
J(\policy)=&\E_{s_0\sim\rho_{0}(\cdot)}[\varphi(s_0)]
+
\dfrac{1}{1-\tilde\gamma}\sum_{s\in\calS}d^{\lambda}_{\policy}(s)
\left(
\sum_{t=0}^{\infty}\gamma^t \lambda^t
\delta^{\varphi}_{\policy,t}(s)
\right)
\\
\nonumber
=&\E_{s_0\sim\rho_{0}(\cdot)}[\varphi(s_0)]
+
\dfrac{1}{1-\tilde\gamma}\E_{s\sim d^{\lambda}_{\policy}(\cdot)}
\left[
\sum_{t=0}^{\infty}\gamma^t \lambda^t
\delta^{\varphi}_{\policy,t}(s)
\right]
.
\end{flalign}
This concludes the proof of Theorem \ref{objective-td-error-version}.
\end{proof}

\subsection{General Discussion}

The objective shown Theorem \ref{objective-td-error-version}
\begin{flalign}
J(\policy)=&\E_{s_0\sim\rho_{0}(\cdot)}[\varphi(s_0)]
+
\dfrac{1}{1-\tilde\gamma}\sum_{s\in\calS}d^{\lambda}_{\policy}(s)
\left(
\sum_{t=0}^{\infty}\gamma^t \lambda^t
\delta^{\varphi}_{\policy,t}(s)
\right)
\\
=&\E_{s_0\sim\rho_{0}(\cdot)}[\varphi(s_0)]
+
\dfrac{1}{1-\tilde\gamma}\E_{s\sim d^{\lambda}_{\policy}(\cdot)}
\left[
\sum_{t=0}^{\infty}\gamma^t \lambda^t
\delta^{\varphi}_{\policy,t}(s)
\right]
\end{flalign}
unifies previous results according to the following way: 
\begin{itemize}
\item if $\varphi(s)=V_{\pi}(s)$, then Theorem \ref{objective-td-error-version} implies the objective shown in Theorem \ref{lem:lam-return-objective}, i.e.,
\begin{flalign}
J({{\pi}})=\dfrac{1}{1-{\tilde{\gamma}}}\sum_{s\in\calS}d^{\lambda}_{{\pi}}(s)R^{(\lambda)}_{\pi}(s)=
\dfrac{1}{1-{\tilde{\gamma}}}\E_{s\sim d^{\lambda}_{{\pi}}(\cdot)}
\left[R^{(\lambda)}_{\pi}(s)\right];
\end{flalign}
\item if $\varphi(s)=V_{\pi}(s)$ and $\lambda\rightarrow0$,  then Theorem \ref{objective-td-error-version} implies the objective shown in Theorem \ref{them:obj-01}, i.e.,
\begin{flalign}
J({\pi})
=&
\dfrac{1}{1-\gamma}\sum_{s_0\in\calS}\rho_{0}(s_0)\sum_{s\in\calS}d^{s_0}_{\pi}(s) R_{{\pi}}(s)\\
=&\dfrac{1}{1-\gamma}\mathbb{E}_{s\sim d_{\pi}^{\rho_0}(\cdot),a\sim\pi(\cdot|s),s^{'}\sim\Pro(\cdot|s,a)}\left[r(s^{'}|s,a)\right].
\end{flalign}
\end{itemize}

\subsection{Application}
\label{application-gae}

In this section, we apply Theorem \ref{objective-td-error-version} to provide an equivalent policy optimization shown in \citep[Section 6.1]{schulman2016high}. This policy optimization formally establishes a problem with respect to GAE, which is widely used in modern reinforcement learning.

\begin{proposition}\emph{(\citep[Proposition 1]{yang2022constrained})}
\label{propo-01}
For any two policies $\pi$ and $\pi^{'}$, let 
\begin{flalign}
\nonumber
&\epsilon^{\varphi}_{\policy,t}=:\max_{s_{t}\in\calS}\left\{\E_{a_t\sim\policy(\cdot|s_t),s_{t+1}\sim\Pro(\cdot|s_t,a_t)}[|\delta_{t}^{\varphi}|]\right\},\\
\nonumber
&\epsilon^{V}_{\policy}(\policyy)=:\sup_{t\in\N^{+}}\{\epsilon^{\varphi}_{\policy,t}: \varphi=V_{\policyy}\},\\
\nonumber
&D_{\mathrm{TV}}(\policy,\policyy)[s]=\dfrac{1}{2}\sum_{a\in\calA}\left|\pi(a|s)-\pi(a^{'}|s)\right|,
\end{flalign}
then
\begin{flalign}
\nonumber
&J(\policy)-J(\policyy)
\ge\frac{1}{1-\tilde\gamma}\E_{s\sim{d}_{\policyy}^{\lambda}(\cdot),a\sim\policy(\cdot|s)}
\Bigg[A^{\mathtt{GAE}(\gamma,\lambda)}_{\policyy}(s,a)\\
\label{pro1-bound-01}
&~~~~~~~~~~~~~~~~~~~~~~~~~~~~~~~~~~~~~~~~~~~~~~~~~~~~~~~~~~-\dfrac{2\tilde{\gamma}\left(\gamma\lambda(|\calS|-1)+1\right)\epsilon^{V}_{\policy}(\policyy)}{(1-\tilde\gamma)(1-\gamma\lambda)}D_{\mathrm{TV}}(\policy,\policyy)[s]\Bigg],
\end{flalign}
where we consider the pair $(s,a)$ stats at time $t$, i,e., $(s,a)=(s_t,a_t)$, and 
 \begin{flalign}
{A}^{\mathtt{GAE}(\gamma,\lambda)}_{\pi^{'}}(s_t,a_t)=\sum_{\ell=0}^{\infty}(\gamma\lambda)^{\ell}\E_{s_{t+\ell+1}}\left[\delta^{V_{\pi^{'}}}_{t+\ell}\right],
\end{flalign}
where $\delta^{V}_{t}=r_{t+1}+\gamma V(s_{t+1})-V(s_{t})$ is TD error, and $V(\cdot)$ is an estimator of value function.
\end{proposition}

Furthermore, we know,
\begin{flalign}
\label{inequlities}
\E_{s\sim{d}_{\policyy}^{\lambda}(\cdot)}\left[D_{\text{TV}}(\policy,\policyy)[s]\right]
\leq&
\E_{s\sim{d}_{\policyy}^{\lambda}(\cdot)}\left[\sqrt{\frac{1}{2}\text{KL}(\policy,\policyy)[s]}\right]
\leq
\sqrt{\frac{1}{2}\E_{s\sim{d}_{\policyy}^{\lambda}(\cdot)}\left[\text{KL}(\policy,\policyy)[s]\right]},
\end{flalign}
where $\text{KL}(\cdot,\cdot)$ is KL-divergence, and \[\text{KL}(\policy,\policyy)[s]=\text{KL}(\policy(\cdot|s),\policyy(\cdot|s));\] the first inequality follows Pinsker's inequality \citep{csiszar2011information} and the second inequality follows Jensen's inequality.
Then the bound shown in (\ref{pro1-bound-01}) holds if we make the following substitution:
\[
\E_{s\sim{d}_{\policyy}^{\lambda}(\cdot)}\left[D_{\mathrm{TV}}(\policy,\policyy)[s]\right]
\leftarrow
\sqrt{\frac{1}{2}\E_{s\sim{d}_{\policyy}^{\lambda}(\cdot)}\left[\mathrm{KL}(\policy,\policyy)[s]\right]}.
\]

Finally, according to trust region methods, we obtain the following policy optimization problem,
\begin{flalign}
\max_{\pi\in\Pi}&\E_{s\sim{d}_{\pi_{k}}^{\lambda}(\cdot),a\sim\policy(\cdot|s)}
\left[A^{\mathtt{GAE}(\gamma,\lambda)}_{\pi_{k}}(s,a)\right],\\
&\mathrm{s.t.,}{\E_{s\sim{d}_{\pi_{k}}^{\lambda}(\cdot)}\left[\mathrm{KL}(\policy,\pi_{k})[s]\right]}\leq \delta,
\end{flalign}
which unifies \citep[Section 6.1]{schulman2016high} and provides a theoretical fundament for policy optimization with GAE.

\section{Bibliographical Remarks}

The objective presented in Theorem \ref{them:obj-01} has been widely used in extensive reinforcement learning literature (e.g., \citep{silver2014deterministic}), which is a fundamental way to understand the stochastic influence of the objective.
Theorem \ref{lem:lam-return-objective} is parallel to Theorem \ref{them:obj-01}, where Theorem \ref{lem:lam-return-objective} considers the $\lambda$-version dynamics.
$\lambda$-return plays a centre way to show Theorem \ref{lem:lam-return-objective}, where the $\lambda$-return and its error-reduction properties is introduced by \citep{watkins1989learning}. 
The property of $\lambda$-Bellman operator is well-documented (e.g., \citep{bertsekas2022abstract}).
Theorem \ref{objective-td-error-version} is a TD$(\lambda)$ version of objective, TD$(\lambda)$ is introduced by \cite{sutton1984temporal,sutton1988learning}.
Off-line TD$(\lambda)$ share a natural structure with GAE (generalized advantage estimation) \citep{schulman2016high}, Theorem \ref{objective-td-error-version} provides a possible way to formulate the GAE and such an idea has been utilized by \citep{yang2022constrained}.

\section{Conclusion}

This lecture presents a general perspective on RL objectives, where we show three versions of objectives.
The first is the standard definition, then we extend it to the $\lambda$-return version, and the final unifies the previous two versions of objectives.
The last version provides a theoretical fundament for policy optimization with GAE.

\bibliographystyle{plainnat}
\bibliography{ref}

\begin{thebibliography}{13}
\providecommand{\natexlab}[1]{#1}
\providecommand{\url}[1]{\texttt{#1}}
\expandafter\ifx\csname urlstyle\endcsname\relax
  \providecommand{\doi}[1]{doi: #1}\else
  \providecommand{\doi}{doi: \begingroup \urlstyle{rm}\Url}\fi

\bibitem[Achiam et~al.(2017)Achiam, Held, Tamar, and
  Abbeel]{achiam2017constrained}
Joshua Achiam, David Held, Aviv Tamar, and Pieter Abbeel.
\newblock Constrained policy optimization.
\newblock In \emph{International conference on machine learning}, pages 22--31.
  PMLR, 2017.

\bibitem[Bertsekas(2022)]{bertsekas2022abstract}
Dimitri Bertsekas.
\newblock \emph{Abstract dynamic programming}.
\newblock Athena Scientific, 2022.

\bibitem[Csisz{\'a}r and K{\"o}rner(2011)]{csiszar2011information}
Imre Csisz{\'a}r and J{\'a}nos K{\"o}rner.
\newblock \emph{Information theory: coding theorems for discrete memoryless
  systems}.
\newblock Cambridge University Press, 2011.

\bibitem[Howard(1960)]{howard1960dynamic}
Ronald~A Howard.
\newblock Dynamic programming and markov processes.
\newblock 1960.

\bibitem[Puterman(2014)]{puterman2014markov}
Martin~L Puterman.
\newblock \emph{Markov decision processes: discrete stochastic dynamic
  programming}.
\newblock John Wiley \& Sons, 2014.

\bibitem[Schulman et~al.(2016)Schulman, Moritz, Levine, Jordan, and
  Abbeel]{schulman2016high}
John Schulman, Philipp Moritz, Sergey Levine, Michael Jordan, and Pieter
  Abbeel.
\newblock High-dimensional continuous control using generalized advantage
  estimation.
\newblock In \emph{International Conference on Learning Representations}, 2016.

\bibitem[Silver et~al.(2014)Silver, Lever, Heess, Degris, Wierstra, and
  Riedmiller]{silver2014deterministic}
David Silver, Guy Lever, Nicolas Heess, Thomas Degris, Daan Wierstra, and
  Martin Riedmiller.
\newblock Deterministic policy gradient algorithms.
\newblock In \emph{International conference on machine learning}, pages
  387--395. PMLR, 2014.

\bibitem[Sutton(1988)]{sutton1988learning}
Richard~S Sutton.
\newblock Learning to predict by the methods of temporal differences.
\newblock \emph{Machine learning}, 3\penalty0 (1):\penalty0 9--44, 1988.

\bibitem[Sutton and Barto(2018)]{sutton2018reinforcement}
Richard~S Sutton and Andrew~G Barto.
\newblock \emph{Reinforcement learning: An introduction}.
\newblock MIT press, 2018.

\bibitem[Sutton(1984)]{sutton1984temporal}
Richard~Stuart Sutton.
\newblock \emph{Temporal credit assignment in reinforcement learning}.
\newblock University of Massachusetts Amherst, 1984.

\bibitem[Watkins(1989)]{watkins1989learning}
Christopher John Cornish~Hellaby Watkins.
\newblock Learning from delayed rewards.
\newblock 1989.

\bibitem[Weng(2018)]{weng2018PG}
Lilian Weng.
\newblock Policy gradient algorithms.
\newblock \emph{lilianweng.github.io/lil-log}, 2018.
\newblock URL
  \url{https://lilianweng.github.io/lil-log/2018/04/08/policy-gradient-algorithms.html}.

\bibitem[Yang et~al.(2022)Yang, Ji, Dai, Zhang, Zhou, Li, Yang, and
  Pan]{yang2022constrained}
Long Yang, Jiaming Ji, Juntao Dai, Linrui Zhang, Binbin Zhou, Pengfei Li,
  Yaodong Yang, and Gang Pan.
\newblock Constrained update projection approach to safe policy optimization.
\newblock In \emph{36th Conference on Neural Information Processing Systems},
  2022.

\end{thebibliography}

\end{document}